%% file: ner.tex
\documentclass[11pt,a4paper]{article}
\usepackage[final]{eacl2021}
\usepackage{times}
\usepackage{latexsym}

\usepackage{microtype}

\usepackage{multirow}
\usepackage{graphicx}
\usepackage{tikz-dependency}
\usepackage{subcaption}

\usepackage{tablefootnote}

\usepackage{tabularx}
\newcolumntype{R}[1]{>{\hsize=#1\hsize\raggedleft\arraybackslash}X}%
\newcolumntype{L}[1]{>{\hsize=#1\hsize\raggedright\arraybackslash}X}%
\newcolumntype{C}[1]{>{\hsize=#1\hsize\centering\arraybackslash}X}%

\usepackage{tikz}
\newcommand\orgbox[2][]{\tikz[overlay]\node[fill=blue!20,inner sep=1pt, anchor=text, rectangle, rounded corners=1mm,#1] {#2};\phantom{#2}}
\newcommand\perbox[2][]{\tikz[overlay]\node[fill=green!20,inner sep=1pt, anchor=text, rectangle, rounded corners=1mm,#1] {#2};\phantom{#2}}
\newcommand\locbox[2][]{\tikz[overlay]\node[fill=red!20,inner sep=1pt, anchor=text, rectangle, rounded corners=1mm,#1] {#2};\phantom{#2}}
\newcommand\miscbox[2][]{\tikz[overlay]\node[fill=yellow!20,inner sep=1pt, anchor=text, rectangle, rounded corners=1mm,#1] {#2};\phantom{#2}}
\newcommand\obox[2][]{\tikz[overlay]\node[fill=gray!20,inner sep=1pt, anchor=text, rectangle, rounded corners=1mm,#1] {#2};\phantom{#2}}

\definecolor{green}{rgb}{0.0, 0.75, 0.0}
	
\title{Entity-Switched Datasets: An Approach to Auditing the In-Domain Robustness of Named Entity Recognition Models}

\author{Oshin Agarwal \\
  University of Pennsylvania \\
  \texttt{oagarwal@seas.upenn.edu} \\\And
  Yinfei Yang \\
  Google AI \\
  \texttt{yinfeiy@google.com} \\\AND
  Byron C. Wallace \\
  Northeastern University \\
  \texttt{b.wallace@northeastern.edu} \\\And
  Ani Nenkova \\
  University of Pennsylvania \\
  \texttt{nenkova@seas.upenn.edu} \\
  }

\date{}

\begin{document}
\maketitle

\begin{abstract}
Named entity recognition (NER) systems perform well on standard datasets comprising English news. But given the paucity of data, it is difficult to draw conclusions about the robustness of systems with respect to recognizing a diverse set of entities. We propose a method for auditing the in-domain robustness of systems, focusing specifically on differences in performance due to the national origin of entities. We create \emph{entity-switched} datasets, in which named entities in the original texts are replaced by plausible named entities of the same type but of different national origin.
We find that state-of-the-art systems' performance vary widely even in-domain: In the same context, entities from certain origins are more reliably recognized than entities from elsewhere. 
This auditing approach can facilitate the development of more robust NER systems, and will allow research in this area to consider fairness criteria that have received heightened attention in other predictive tasks.\footnote{
The datasets are available at \href{https://github.com/oagarwal/entity-switched-ner}{https://github.com/oagarwal/entity-switched-ner}. 
Our datasets are still limited to top few populous countries. However, they can easily be created for more rare entities as well. Code for the same will be released.} 

\end{abstract}


\section{Introduction}

Named Entity Recognition (NER) systems work well on domains such as English news, achieving high performance on standard datasets such as MUC-6 \cite{grishman1996message}, CoNLL 2003 \cite{tjong-kim-sang-de-meulder-2003-introduction} and OntoNotes \cite{pradhan-xue-2009-ontonotes}. Research in other areas of predictive technology has revealed, however, that ostensibly strong predictive performance may obscure wide variation in performance for certain types of data. For example, gender recognition systems attain high accuracy on what used to be considered standard datasets for this task, but have large error rates on people with dark skin tone, particularly on women with dark skin \cite{DBLP:conf/fat/BuolamwiniG18}. Language identification is also highly accurate on standard datasets \cite{DBLP:conf/emnlp/ZhangRGBBW18} but may fail to recognize dialects, e.g., failing to identifying African American English as English \cite{DBLP:conf/emnlp/BlodgettGO16}.

\begin{table*}[t]
\centering
\small
\setlength{\tabcolsep}{2pt}
\begin{tabularx}{0.9\linewidth}{L{0.5}|L{0.5}}
\multicolumn{1}{c|}{\bf Original Sentence} & \multicolumn{1}{c}{\bf New Sentence} \\
\hline
Defender \perbox{Hassan Abbas {\bf \tiny PER}} rose to intercept a ... & Defender \perbox{Ritwika Tomar {\bf \tiny PER}} rose to intercept a ... \\\hline
The Democratic Convention signed an agreement on government and parliamentary support with its coalition partners the \orgbox{Social Democratic Union {\bf \tiny ORG}} and the\orgbox{Hungarian Democratic Union {\bf \tiny ORG}}. & The Democratic Convention signed an agreement on government and parliamentary support with its coalition partners the \orgbox{Jharkhand Mukti Morcha {\bf \tiny ORG}} and the \orgbox{Mizo National Front {\bf \tiny ORG}}. \\
\end{tabularx} 
\caption{Example of switching entities.}
\label{table:SwitchExample}
\end{table*}

Here, we set out to develop methods for testing two intertwined properties of NER models: (i) Their robustness on a variety of entities, and, (ii) their relative performance across groups, which here correspond to national origin. It is known that the robustness of NER methods depends on entities being represented in the training data \cite{augenstein2017generalisation}. Probing their performance through the lens of national origin is then one way to question the choice of training data, and what a system can learn from it. This has important fairness implications: Our approach enables auditing of the systems for entity groups, e.g., ethnic groups within a country of origin, in line with guidelines for model card reporting of system strengths and weaknesses \cite{DBLP:conf/fat/MitchellWZBVHSR19}. 

We propose a diagnostic to evaluate the in-domain robustness of NER models by expanding existing datasets programmatically. These generated datasets not only have a diverse set of entities but also the same entity in a variety of contexts. This will allow for a more robust and large scale evaluation of NER. We evaluate state-of-the-art systems on these \emph{entity-switched} datasets and find that they have highest performance (F1) on American and Indian entities, and lowest performance on Vietnamese and Indonesian entities.

\section{Entity-Switched Datasets}

To create entity-switched datasets, we replace entities in existing datasets with names from various countries while retaining the rest of the text and maintaining its coherence. An example is shown in Table \ref{table:SwitchExample} were the original entities are replaced by Indian ones. While some types of entities (such as persons) can be readily replaced, other entities (such as organizations) require more care to maintain the coherence of the text. For this reason, we use two versions of the datasets. In one, we replace all entities; in the other, we replace only \emph{PER}. Perturbation techniques have also been used in \cite{prabhakaran-etal-2019-perturbation} to study the impact of person names on toxicity in online comments by substituting in names of controversial personalities, in \cite{emami-etal-2019-knowref} to make coreference resolution robust to gender and number cues by making both antecedent and candidate of the same type, in \cite{subramanian2019improving} for adverserial training for coreference resolution of person and location entities, in \cite{raiman-miller-2017-globally} to augment question-answering datasets and in \cite{kaushik2019learning} to create manually perturbed examples by experts for sentiment analysis and natural language inference. However, there is no large scale dataset available for the evaluation of NER on ethnically diverse entities.

\input{conll_names.tex}

\input{ontonotes_names.tex}

\input{top_names.tex}


\input{conll_entities.tex}

\subsection{Replacing PER entities}

In the first group of datasets, we change only names of people. We replace all \emph{PER} entities in the test set with the same string. For example, all sequences of \emph{PER} might be replaced with the name `John Smith', or the name `Marijuana Pepsi'.\footnote{\href{https://tinyurl.com/yxdhupr9}{https://tinyurl.com/yxdhupr9}} Names for replacement are drawn from lists of popular names in the countries with the largest populations. This allows us to examine system performance with respect to country of origin, and also in terms of the number of people whose names would be potentially affected by recognition failures. Specifically, we selected the 15 most populous countries and found 20 common first names and 10 common family names for each.\footnote{Sources include: Wikipedia, websites with baby names, and websites listing popular names} We used two sets of Chinese names, from mainland China and Taiwan, respectively. We also use two sets of U.S. names: The first comprising common names (e.g., John Smith) and the second composed of Native American, African American, and names that could be locations (e.g., Madison) or regular words (e.g., Brown). 

The first names used in the experiment are a mix of male, female, and unisex names. We create full names by matching each first name to a random family name. For some countries, names have additional constraints that we account for. In Indonesia,\footnote{\href{https://en.wikipedia.org/wiki/Indonesian\_names}{https://en.wikipedia.org/wiki/Indonesian\_names}} names might include only a single name or multiple first names (without a family name). In Pakistan,\footnote{\href{https://en.wikipedia.org/wiki/Pakistani\_name}{https://en.wikipedia.org/wiki/Pakistani\_name}} some female names have a first name followed by the father/husband's most called name.

We replace all \emph{PER} spans in the test set with a single name.\footnote{Training data is unchanged, as our goal is to quantify the robustness of identifying various names.} We attempt to be consistent in the replacement, i.e., full names are replaced by full names, first names by first names, and last names by last names. We treat space-separated multi-word names as full names. We take a Western-centric view and consider the first word to be a first name and the remaining to be the last name, and determine if other occurrences are first or last names by string matching. If a multi-word name is part of a longer name, we do not break it down and replace it based on the longer name it matches. 

We use the English CoNLL'03 and calculate the F1 for each country, replacing every \emph{PER} entity in the test set with each of the 20 names in turn and taking the average over the 20 versions of the dataset. We evaluate the word-based biLSTM-CRF \cite{huang2015bidirectional}, word and chararacter-based biLSTM-CRF \cite{lample-etal-2016-neural} and BERT \cite{devlin-etal-2019-bert}. For the first two, we used 300-d cased GloVe \cite{pennington2014glove} vectors trained on Common Crawl.\footnote{http://nlp.stanford.edu/data/glove.840B.zip} For BERT, we use the public large
 uncased \footnote{Uncased performed better than cased.} model and apply the default fine-tuning strategy. We use IO labeling and evaluate systems using micro-F1 at the token level. 

We present the results in Table \ref{table:NamesCoNLL}. All the models achieve higher F1 on typical American names, Russian, Indian and Mexican names than the original dataset ({\em Super recall}). Precision remains the same but recall improves to almost perfect. 

For the GloVe models, performance drops by up to $\sim$10 points F1 for certain countries ({\em Poor recall}). Names from Indonesia and Vietnam fare the worst, along with the difficult US names and names from Taiwan, with small degradation of precision and a precipitous drop of recall. BERT exhibits a similar pattern, with stable precision and varying recall which remains above 84\% for all name origins. The names with the highest and lowest F1 with \cite{huang2015bidirectional} are shown in Table \ref{table:TopNames}.

Notably, BERT performance is lowest on names from Ethiopia, Nigeria, and the Philippines (see {\em BERT not best} rows). In light of these findings, one might wonder if accepting current architectures trained on standard corpora as state-of-the-art is the NER equivalent of developments in photography, which was optimized for perfect exposure of white skin, and which is the assumed technical reason for many failures of computer vision applications when applied to dark skinned people \cite{RaceAfterTech}. At the very least practitioners should be cognizant of these systematic performance differences.

BERT and character LSTM-CRF results are higher and more stable, but it is nevertheless clear that one need not perform a completely out-of-domain test to observe deteriorating performance; changing the name strings is sufficient. We perform similar tests on the newswire section of OntoNotes. We use the original train and test splits, where we have switched \emph{PER} in the latter. We use names from India and Vietnam because these were in the top and bottom performing entity-switched sets, respectively, for CoNLL data. We observe a similar drop in performance (Table \ref{table:NamesOntonotes}). We hypothesize that this is due to the names that systems have seen during pre-training and training. Based on a similar hypothesis, \citet{derczynski-etal-2017-results} collected a dataset with no overlap in train and test set entities.

\subsection{Replacing other entity types}

We construct two more datasets derived from the original English CoNLL test set. In these we replace all \emph{PER}, \emph{LOC} and \emph{ORG} instances with entities of the same type and from a particular country. We do not replace \emph{MISC} entities, because these are not usually country-specific. In one dataset, we replace all entities with corresponding entities of Indian origin; in the other, with entities of Vietnamese origin. Unlike in the above dataset, where we replaced every entity of type \emph{PER} with the \emph{same} name throughout, we perform a stochastic (though type-constrained) mapping from the original set of entities to the new entity set. That is, when replacing a target entity $e$ of type $t$, we sample an entity with which to replace it at i.i.d. random from entities of type $t$ in the set of country-specific entities. 

We select \emph{PER} names as in the previous section. For each document, we then generate a list of possible names an entity was referred to by string matching and replace each entity with a new name as consistently as possible, as in the previous section. For \emph{LOC}, we select names of villages, cities, and provinces and again select a location of the same type from the country-specific list in a document.

Consistently replacing \emph{ORG} entities is more complicated than replacing \emph{PER} and \emph{LOC} entities, because not every organization would be suitable for every context. We cannot, for instance, reasonably replace `Bank of America' in `We withdrew money from the Bank of America' with `New York Times' or `Mayo Clinic'. For this reason, we divided organizations into sub-categories, selected candidates for each category from similar websites as above, labelled test \emph{ORG} with the sub-category manually, and then replaced them a country-specific \emph{ORG} of the same sub-category, again being consistent within a document based on string matching. The sub-categories we used are: Airline, Bank, Corporation, Newspaper, Political Party, Restaurant, Sports Team, Sports Union, University, Others. Others included international or intergovernmental organizations such as United Nations and we did not replace these. 

We observe a drop in performance on both the datasets (Table \ref{table:EntitiesCoNLL}). 
Both precision and recall drop for the GloVe systems. However, for BERT, the precision remains the same and only the recall drops. 

\input{examples}

\subsection{Error Analysis}

We show some examples of the replacement and predictions by \cite{huang2015bidirectional} in Table \ref{table:Examples}. They show that systems likely ignore predictive contexts and assign labels based mostly on the word identity. In row 1, both the entities should be of the same type -- \emph{LOC} or \emph{PER}; however for Indian entities, one is predicted as \emph{PER} and one as \emph{LOC}; for Vietnamese, one is predicted as \emph{PER} and the other isn't even recognized as an entity. Row 2 is indeed hard and a known word identity would be easier to detect in this case. Row 3 is a common pattern `\emph{LOC} \emph{DATE}' from the training data but the systems do not recognize these for the switched datasets. In fact, for Vietnamese it even causes the date to be predicted as a \emph{LOC}. Similarly, row 6 is another common pattern in the training data with sports scores but the system fails to recognize it. In rows 5 and 7, there are clear contextual clues for the entity type but the system fails to consider it for prediction. In row 5, anything following `newspaper' is the newspaper's name and should be \emph{ORG}. Similarly in row 7, anything following `team' is the team's name and should be \emph{ORG}. While it may appear the `Italian newspaper' followed by a name from a different origin may the cause of the error, in the last row all the names followed by German are recognized correctly irrespective of origin and it doesn't seem that systems learns this level of knowledge. 

\section{Conclusion}
\label{sec:future}
Standard NER datasets such as English news contain a limited number of unique entities, with many of them occurring in both the train and the test sets. State-of-the-art models may thus memorize observed names, rather than learning to identify entities on the basis of context. As a result, models perform less well on `foreign' entity instances.

To measure this, we introduced a practical approach of \emph{entity-switching} to create datasets to test the in-domain robustness of systems. We selected entities from different countries and showed that models perform extremely well on entities from certain countries, but not as well on others. This finding has fairness implications when NER is used in practice. We hope that these datasets 
will facilitate research in developing more robust systems. 


\bibliographystyle{acl_natbib}
\bibliography{ner}

\end{document}

%% file: conll_names.tex
\begin{table*}[t]
\centering
\small
\setlength{\tabcolsep}{4pt}
\begin{tabularx}{\linewidth}{L{0.6}|C{0.2}C{0.2}C{0.2}|C{0.2}C{0.2}C{0.2}|C{0.2}C{0.2}C{0.2}}
\cline{2-10}
 & \multicolumn{3}{c|}{\cite{huang2015bidirectional}} & \multicolumn{3}{c|}{\cite{lample-etal-2016-neural}} & \multicolumn{3}{c}{\cite{devlin-etal-2019-bert}}\\
& \multicolumn{3}{c|} {\scriptsize GloVe words} & \multicolumn{3}{c|} {\scriptsize GloVe words+chars} & \multicolumn{3}{c} {\scriptsize BERT subwords}\\
\cline{2-10}
&P&R&F1&P&R&F1&P&R&F1\\\hline
\em{Original testset} & 96.9 & 96.5 & 96.7 & 97.1 & 98.1 & 97.6 & 98.3 & 98.1 & 98.2\\ \hline \hline
\em{Super recall} & & & & & & & & & \\
US & 96.9 & 99.6 & 98.2 & 96.9 & 99.6 & 98.3 & 98.4 & 99.7 & 99.1\\
Russia & 96.8 & 99.5 & 98.1 & 97.1 & 99.8 & 98.4 & 98.4 & 99.3 & 98.9\\
India & 96.5 & 99.5 & 98.0 & 97.1 & 99.3 & 98.2 & 98.4 & 98.8 & 98.6\\
Mexico & 96.7 & 98.9 & 97.8 & 97.1 & 98.9 & 98.0 & 98.4 & 99.2 & 98.8\\ \hline \hline
\em{Poor recall} & & & & & & & & & \\
China-Taiwan & 95.4 & 93.2 & 93.9 & 97.0 & 94.9 & 95.6 & 98.3 & 92.0 & 94.8\\
US (Difficult) & 95.9 & 87.4 & 90.2 & 96.6 & 87.9 & 90.7 & 98.1 & 88.5 & 92.3\\
Indonesia & 95.3 & 84.6 & 88.7 & 96.5 & 91.0 & 93.3 & 97.8 & 85.8 & 92.0\\
Vietnam & 94.6 & 78.2 & 84.2 & 96.0 & 78.5 & 84.5 & 98.0 & 84.2 & 89.8 \\ \hline \hline
\em{BERT not best} & & & & & & & & & \\ 
Ethiopia & 96.5 & 96.8 & 96.6 & 96.6 & 98.6 & 97.9 & 98.3 & 90.6 & 94.1 \\
Nigeria & 96.3 & 92.2 & 94.1 & 97.1 & 96.6 & 96.8 & 98.2 & 90.2 & 93.8 \\
Philippines & 97.3 & 97.9 & 97.5 & 97.5 & 98.9 & 98.2 & 98.6 & 94.7 & 96.4 \\ \hline \hline
\em{Other} & & & & & & & & & \\
Bangladesh & 96.7 & 97.5 & 97.1 & 97.1 & 97.6 & 97.3 & 98.4 & 97.8 & 98.0 \\
Brazil & 96.6 & 96.8 & 96.6 & 97.1 & 96.2 & 96.5 & 98.4 & 96.7 & 97.5 \\
China-Mainland & 95.7 & 97.9 & 96.7 & 97.0 & 97.4 & 97.2 & 98.4 & 96.7 & 97.5 \\
Egypt & 96.6 & 99.2 & 97.8 & 97.0 & 98.2 & 97.6 & 98.4 & 97.4 & 97.9 \\
Japan & 96.7 & 97.2 & 96.8 & 97.0 & 98.7 & 97.8 & 98.5 & 99.0 & 98.7 \\
Pakistan & 96.2 & 92.6 & 94.1 & 97.0 & 96.5 & 96.6 & 98.3 & 95.3 & 96.7 \\
\end{tabularx} 
\caption{Performance of systems on PER entity of CoNLL 03 test data. Original refers to the unchanged data. The rest of the rows are averages over 20 names typical for each country.}\label{table:NamesCoNLL}
\vspace{-1em}
\end{table*}

%% file: ontonotes_names.tex
\begin{table*}[t]
\centering
\small
\setlength{\tabcolsep}{4pt}
\begin{tabularx}{\linewidth}{L{0.6}|C{0.2}C{0.2}C{0.2}|C{0.2}C{0.2}C{0.2}|C{0.2}C{0.2}C{0.2}}
\cline{2-10}
& \multicolumn{3}{c|}{\cite{huang2015bidirectional}} & \multicolumn{3}{c|}{\cite{lample-etal-2016-neural}} & \multicolumn{3}{c}{\cite{devlin-etal-2019-bert}}\\
& \multicolumn{3}{c|} {\scriptsize GloVe words} & \multicolumn{3}{c|} {\scriptsize GloVe words+chars} & \multicolumn{3}{c} {\scriptsize BERT subwords}\\
\cline{2-10}
&P&R&F1&P&R&F1&P&R&F1\\\hline
Original & 94.7 & 95.6 & 95.2 & 97.5 & 95.0 & 96.2 & 97.0 & 96.8 & 96.9\\ \hline
India    & 94.2 & 95.5 & 94.8 & 97.0 & 95.7 & 96.2 & 96.3 & 96.9 & 96.6\\
Vietnam  & 93.1 & 82.3 & 85.8 & 96.3 & 82.3 & 86.9 & 96.5 & 85.2 & 90.5\\
\end{tabularx} 
\caption{Performance of systems on PER entity of OntoNotes newswire test data. Original refers to the unchanged data. The rest of the rows report averages over 20 names typical for each country.}\label{table:NamesOntonotes}
\end{table*}

%% file: top_names.tex
\begin{table*}[t]
\centering
\footnotesize
\setlength{\tabcolsep}{2pt}
\begin{tabularx}{0.75\linewidth}{L{0.6}L{0.5}C{0.2}|L{0.6}L{0.5}C{0.2}}
\hline
\multicolumn{3}{c|}{\bf{Highest performance}}&\multicolumn{3}{c}{\bf{Lowest performance}}\\
Name & Country & F1 & Name & Country & F1\\\hline
Jose Mari Andrada & Philippines & 98.8 & Trinity Washington & U.S.(Difficult) & 37.8\\
Chris Collins & U.S. & 98.4 & My On & Vietnam & 37.9\\
Alex Mikhailov & Russia & 98.4 & Thien Thai & Vietnam & 62.5\\
Alejandro Garcia & Mexico & 98.4 & Elaf Zahaar & Pakistan & 69.9\\
\end{tabularx} 
\caption{Names on which \cite{huang2015bidirectional} achieved highest and lowest F1 scores.}\label{table:TopNames}
\end{table*}

%% file: conll_entities.tex
\begin{table*}[t]
\centering
\small
\setlength{\tabcolsep}{4pt}
\begin{tabularx}{\linewidth}{L{0.6}|C{0.2}C{0.2}C{0.2}|C{0.2}C{0.2}C{0.2}|C{0.2}C{0.2}C{0.2}}
\cline{2-10}
& \multicolumn{3}{c|}{\cite{huang2015bidirectional}} & \multicolumn{3}{c|}{\cite{lample-etal-2016-neural}} & \multicolumn{3}{c}{\cite{devlin-etal-2019-bert}}\\
& \multicolumn{3}{c|} {\scriptsize GloVe words} & \multicolumn{3}{c|} {\scriptsize GloVe words+chars} & \multicolumn{3}{c} {\scriptsize BERT subwords}\\
\cline{2-10}
&P&R&F1&P&R&F1&P&R&F1\\\hline
Original & 90.9 & 91.4 & 91.2 & 90.9 & 92.6 & 91.7 & 95.5 & 93.3 & 94.4\\ \hline
India & 84.3 & 77.3 & 80.7 & 83.8 & 82.9 & 83.3 & 95.6 & 87.8 & 91.5 \\
Vietnam & 74.3 & 73.0 & 73.6 & 77.9 & 76.4 & 77.2 & 96.2 & 81.5 & 88.2 \\
\end{tabularx} 
\caption{Performance of systems on all entity types in the CoNLL 03 test data.}\label{table:EntitiesCoNLL}
\end{table*}

%% file: examples.tex
\begin{table*}[t]
\centering
\small
\setlength{\tabcolsep}{2pt}
\begin{tabularx}{\linewidth}{L{0.5}L{0.5}L{0.5}L{0.5}}
{\bf True labels} & {\bf Original} & {\bf Indian} & {\bf Vietnamese} \\\hline
\locbox{Japan {\bf \tiny LOC}} began the defence of their \miscbox{Asian Cup {\bf \tiny MISC}} title with a lucky 2-1 win against \locbox{Syria {\bf \tiny LOC}} in a Group C championship match on Friday. & 
\locbox{Japan {\bf \tiny LOC}} began the defence of their \miscbox{Asian Cup {\bf \tiny MISC}} title with a lucky 2-1 win against \locbox{Syria {\bf \tiny LOC}} in a Group C championship match on Friday. & 
\locbox{Dhanbad {\bf \tiny LOC}} began the defence of their \miscbox{Asian Cup {\bf \tiny MISC}} title with a lucky 2-1 win against \perbox{Thungapuram {\bf \tiny PER}} in a Group C championship match on Friday. & 
\obox{Long An {\bf \tiny O}} began the defence of their \miscbox{Asian Cup {\bf \tiny MISC}} title with a lucky 2-1 win against \perbox{Bac Lieu {\bf \tiny PER}} in a Group C championship match on Friday. \\\hline

\perbox{Nader Jokhadar {\bf \tiny PER}} had given \locbox{Syria {\bf \tiny LOC}} the lead with a well-struck header in the seventh minute. &
\perbox{Nader Jokhadar {\bf \tiny PER}} had given \locbox{Syria {\bf \tiny LOC}} the lead with a well-struck header in the seventh minute. &
\locbox{Priya Khemka {\bf \tiny LOC}} had given \obox{Thungapuram {\bf \tiny O}} the lead with a well-struck header in the seventh minute. &
\locbox{Thien Hue {\bf \tiny LOC}} had given \perbox{Bac Lieu {\bf \tiny PER}} the lead with a well-struck header in the seventh minute. \\\hline

\locbox{ROME {\bf \tiny LOC}} 1996-12-06 & 
\locbox{ROME {\bf \tiny LOC}} 1996-12-06 & 
\obox{Thevaiyur {\bf \tiny O}} 1996-12-06 & 
\locbox{Ha Long Bay 1996-12-06 {\bf \tiny LOC}} \\\hline

SOCCER - \orgbox{FIFA {\bf \tiny ORG}} BOSS \perbox{HAVELANGE {\bf \tiny PER}} STANDS BY \perbox{WEAH {\bf \tiny PER}}. &
SOCCER - \orgbox{FIFA {\bf \tiny ORG}} BOSS \perbox{HAVELANGE {\bf \tiny PER}} STANDS BY \obox{WEAH {\bf \tiny O}}. &
SOCCER - \orgbox{Judo Federation {\bf \tiny ORG}} \obox{of {\bf \tiny O}} \locbox{India {\bf \tiny LOC}} BOSS \perbox{Dheeraj Uniyal {\bf \tiny PER}} STANDS BY \perbox{Anjali Lal {\bf \tiny PER}}. &
SOCCER - \locbox{Vietnam Football Federation {\bf \tiny LOC}} BOSS \obox{Thu {\bf \tiny O}} \miscbox{Thai {\bf \tiny MISC}} STANDS BY \perbox{Duc Tan {\bf \tiny PER}}. \\\hline

In an interview with the \miscbox{Italian {\bf \tiny MISC}} newspaper \orgbox{Gazzetta dello Sport {\bf \tiny ORG}}, he was quoted as saying \perbox{Weah {\bf \tiny PER}} had been provoked into the assault which left \perbox{Costa {\bf \tiny PER}} with a broken nose. &
In an interview with the \miscbox{Italian {\bf \tiny MISC}} newspaper \orgbox{Gazzetta dello Sport {\bf \tiny ORG}}, he was quoted as saying \perbox{Weah {\bf \tiny PER}} had been provoked into the assault which left \locbox{Costa {\bf \tiny LOC}} with a broken nose. &
In an interview with the \miscbox{Italian {\bf \tiny MISC}} newspaper \perbox{Hari Bhoomi {\bf \tiny PER}}, he was quoted as saying \perbox{Masih {\bf \tiny PER}} had been provoked into the assault which left \obox{Damerla {\bf \tiny O}} with a broken nose. & 
In an interview with the \miscbox{Italian {\bf \tiny MISC}} newspaper \perbox{Tien Phong {\bf \tiny PER}}, he was quoted as saying \perbox{Hue {\bf \tiny PER}} had been provoked into the assault which left \locbox{Giang {\bf \tiny LOC}} with a broken nose. \\\hline

\orgbox{Cagliari {\bf \tiny ORG}} (16) v \orgbox{Reggiana {\bf \tiny ORG}} (18) 1530 &
\orgbox{Cagliari {\bf \tiny ORG}} (16) v \orgbox{Reggiana {\bf \tiny ORG}} (18) 1530 &
\locbox{Thiruvananthapuram {\bf \tiny LOC}} (16) v \locbox{Hyderabad {\bf \tiny LOC}} (18) 1530 & 
\perbox{Sanna Khanh Hoa Futsal Club {\bf \tiny PER}} (16) v \orgbox{Ha Long Bay {\bf \tiny ORG}} (18) 1530 \\\hline

Bottom team \orgbox{Reggiana {\bf \tiny ORG}} are also without a suspended defender, \miscbox{German {\bf \tiny MISC}} \perbox{Dietmar Beiersdorfer {\bf \tiny PER}}. &
Bottom team \orgbox{Reggiana {\bf \tiny ORG}} are also without a suspended defender, \miscbox{German {\bf \tiny MISC}} \perbox{Dietmar Beiersdorfer {\bf \tiny PER}}. &
Bottom team \locbox{Hyderabad {\bf \tiny LOC}} are also without a suspended defender, \miscbox{German {\bf \tiny MISC}} \perbox{Navneet Nedungadi {\bf \tiny PER}}. & 
Bottom team \perbox{Ha Long Bay {\bf \tiny PER}} are also without a suspended defender, \miscbox{German {\bf \tiny MISC}} \perbox{Linh On {\bf \tiny PER}}. \\

\end{tabularx} 
\caption{Predictions of \cite{huang2015bidirectional} on the original, Indian and Vietnamese test sets. Many errors are made on the entity-switched datasets, even in the presence of strong contextual clues, including patterns common in the training data.}\label{table:Examples}
\end{table*}